\documentclass[journal]{IEEEtran}

\usepackage{amsthm} 
\usepackage{xcolor}
\usepackage{graphicx}       % 支持 \resizebox
\usepackage{subcaption}
\usepackage[percent]{overpic}
\usepackage{amsmath,amssymb,amsthm}
\usepackage{thmtools}   
\usepackage{multicol}
 \usepackage{booktabs}
  \usepackage{multirow}
\declaretheoremstyle[
  headfont=\bfseries,
  bodyfont=\itshape,
  spaceabove=8pt, spacebelow=8pt,
  headpunct={\;}, postheadspace=0.5em,
]{cnplain}
\declaretheorem[style=cnplain,name=Definition]{definition}

\usepackage{algorithm}
\usepackage{algorithmic}

\begin{document}

\title{Cheeger--Hodge Contrastive Learning for Structurally Robust Graph Representation Learning}
\author{
    Mengyang~Zhao$^{\dagger}$, 
    Longlong~Li$^{\dagger}$, 
    and~Cunquan~Qu$^{*}$
    
    \thanks{$^{\dagger}$Mengyang~Zhao and Longlong Li contributed equally to this work.}
    
    \thanks{Mengyang~Zhao is with the School of Mathematics and the Data Science Institute, Shandong University, Jinan 250100, China (email: mengyangzhao@mail.sdu.edu.cn).}
    
    \thanks{Longlong Li is with the School of Physical and Mathematical Sciences, Nanyang Technological University, Singapore 637371, Singapore (e-mail: longlong.li@ntu.edu.sg).}
    
    \thanks{$^{*}$Cunquan Qu is the corresponding author and is with the Data Science Institute, Shandong University, Jinan 250100, China (e-mail: cqqu@sdu.edu.cn).}
}

\markboth{Journal of \LaTeX\ Class Files,~Vol.~18, No.~9, September~2020}%
{Shell \MakeLowercase{\textit{et al.}}: Cheeger--Hodge Contrastive Learning}
% Cheeger--Hodge
\maketitle

\begin{abstract}

Graph Contrastive Learning (GCL) has emerged as a prominent framework for unsupervised graph representation learning. However, relying on augmentation design alone to define the invariances learned by GCL can be brittle under structural perturbations. To address this issue, we propose Cheeger--Hodge Contrastive Learning (CHCL), a framework that aligns a perturbation-stable Cheeger--Hodge joint signature across augmented views for robust graph representation learning. The proposed signature combines a Cheeger-inspired connectivity signature derived from the algebraic connectivity \(\lambda_2\) with the low-frequency spectrum of the 1-Hodge Laplacian, thereby capturing both global connectivity and higher-order structural information. By aligning encoder representations with the proposed Cheeger--Hodge joint signature across augmented views, CHCL learns graph embeddings that are robust to local structural perturbations. Extensive experiments on standard benchmarks, transfer settings demonstrate that CHCL consistently improves performance, robustness, and generalization.
\end{abstract}


\IEEEpeerreviewmaketitle

\section{Introduction}

\IEEEPARstart{I}{n} recent years, Graph Neural Networks (GNNs) have shown promising results in applications such as social network analysis~\cite{kipf2017semi, li2025multi}, molecular property prediction~\cite{gilmer2017neural, li2024path}, and protein structure-function analysis~\cite{jumper2021highly, chen2025mrhgnn}. However, these successes often rely on large amounts of high-quality labeled data~\cite{xu2018how, velickovic2018graph, ying2018hierarchical}. In real-world scenarios, such labels are costly to obtain, time-consuming to annotate, and often subject to privacy restrictions~\cite{hu2020pretraining, zitnik2018prioritizing}. To reduce this reliance, unsupervised graph representation learning has emerged as a promising direction, aiming to learn generalizable representations from graph structures and node attributes without supervision~\cite{wu2022graph, hu2024comprehensive, wu2020comprehensive, chen2020imple, yang2024multi}. Among these approaches, GCL has become a representative paradigm: it constructs augmented views of the same graph and learns view-consistent representations in the latent space, often producing discriminative and transferable graph embeddings~\cite{li2022graph, li2022let, you2020graph, yuan2023clustering}.

Despite its empirical success, GCL typically learns explicit invariance induced by data augmentation~\cite{zhang2023spectral, wang2019implicit, tan2024community, zhang2024motif}. In practice, however, common augmentation strategies such as edge dropping and feature masking can inadvertently distort global structural properties even under small local perturbations~\cite{zhao2023unsupervised, peng2024label, zhang2023multi, zhu2023cross, bu2024improving}. This suggests a more principled perspective: rather than relying solely on augmentation design to define what should be preserved, we argue for introducing a graph-intrinsic structural target that remains stable under perturbation and directly guides the contrastive objective, with augmentation serving primarily as a view generation mechanism. Recent efforts have sought to alleviate related limitations by incorporating semantic-aware objectives~\cite{liu2024unbiased, yan2025have, xu2025semantic, liu2024fine}, spectral perturbation modeling~\cite{bao2024co}, and higher-order topological priors~\cite{chen2024topogcl}. However, these approaches typically incorporate only partial structural priors in isolation—such as semantic cues, spectral perturbation information, or higher-order topology—and therefore do not provide a unified, graph-intrinsic, and theoretically grounded structural consistency target for contrastive learning. Meanwhile, most GNN backbones remain dominated by local message passing and neighborhood aggregation, which are effective for modeling local geometry and proximity relations, but are often less effective at capturing long-range dependencies and global structural signals~\cite{zhang2025rhomboid, alon2021on, Nguyen}. This further limits the ability of GCL to explicitly preserve the structural information most relevant to robust representation learning.

To address these limitations, we propose Cheeger--Hodge Contrastive Learning (CHCL), a graph contrastive learning framework built around a graph-intrinsic structural signature as a perturbation-stable consistency target. We argue that robust graph representations should reflect two complementary dimensions of graph structure beyond local neighborhoods: (1) \textit{global connectivity}, which captures bottleneck and community structure at the graph level; and (2) \textit{higher-order topology}, which captures cycle-related and flow-related patterns in the edge space. Concretely, we derive a Cheeger-inspired connectivity signature from the algebraic connectivity \(\lambda_2\) of the normalized graph Laplacian, which serves as a computable spectral proxy for global bottleneck structure through the classical Cheeger inequality~\cite{chung1997spectral, jost2022cheeger}. We further combine this with the low-frequency spectrum of the 1-Hodge Laplacian, which characterizes higher-order topological structure beyond pairwise edges~\cite{grande2024disentangling}. Together, these components form a unified Cheeger--Hodge joint signature that captures complementary structural information at different scales.

Building on this joint signature, CHCL incorporates it as an explicit structural consistency target within the contrastive objective, so that the learned embeddings remain aligned with graph-intrinsic structural properties across augmented views. In this way, CHCL reduces dependence on augmentation heuristics, mitigates local structural bias in GNN backbones, and encourages the learned representation space to preserve both global connectivity and higher-order topological semantics. Figure~\ref{fig:framework} illustrates the overall framework, highlighting how the proposed Cheeger--Hodge joint signature serves as a graph-intrinsic, perturbation-stable consistency target for contrastive alignment across augmented views.

The main contributions of this work are summarized as follows:
\begin{itemize}
    \item We define a Cheeger--Hodge joint signature that unifies a Cheeger-inspired connectivity signature derived from \(\lambda_2\) with the low-frequency spectrum of the 1-Hodge Laplacian.
    \item We propose CHCL, a graph contrastive learning framework that incorporates this signature as a perturbation-stable structural consistency target, enabling joint modeling of local geometric information, global connectivity, and higher-order topological structure within a unified contrastive framework.
    \item We conduct extensive experiments across multiple graph domains and learning settings demonstrating that CHCL consistently outperforms strong baselines.
\end{itemize}

\section{Related Work}

\subsection{Graph Contrastive Learning}

GCL has emerged as a powerful approach for unsupervised graph representation learning. Early methods, such as DGI \cite{velickovic2018deep} and InfoGraph \cite{sun2019infograph}, leveraged mutual information maximization between local node representations and global graph summaries to learn effective graph embeddings. Building on this, GraphCL \cite{you2020graph} introduced a framework where two augmented views of the same graph are contrasted, improving representation quality through diverse augmentation strategies. Subsequent studies have advanced GCL mainly from adaptive augmentation and structure-aware  perspectives. From the perspective of adaptive augmentation, GCA improves view generation by adaptively preserving structurally important components during augmentation \cite{zhu2021graph}, while AutoGCL and related methods learn augmentation policies directly from data rather than relying on fixed heuristics \cite{yin2022autogcl}. From the perspective of structure-aware contrastive learning, CI-GCL and CTAug incorporate semantic or global structural priors, such as community organization and cohesive subgraph patterns, into the contrastive process \cite{tan2024community, wu2024graph}. More recently, structure-aware GCL has moved toward intrinsic global descriptors derived from graph spectra, where spectral signals are incorporated into augmentation or optimization to improve structural discriminability\cite{lin2023spectral, ghose2023spectral, liu2022revisiting}. This line of research is further extended by TopoGCL, which introduces persistent homology into graph contrastive learning and shows that enforcing topological consistency across augmented views can improve robustness and graph-level representation quality \cite{chen2024topogcl}.
Building on the aforementioned research, we propose using the Cheeger--Hodge joint signature as semantic invariance anchors for the graph, applying invariance learning to the graph's overall connectivity and higher-order topological structures. By incorporating these signatures into the optimization objective, the model enables the model to preserve key global structural signals under random augmentation perturbations, thereby reducing sensitivity to augmentation types and perturbation intensities, and improving robust generalization performance across datasets and noisy scenarios.

\subsection{Spectral Characterization of Graph}
Spectral graph theory provides a principled lens for understanding the intrinsic structural properties of graphs through the spectra of Laplacian operators. Classical results such as the Cheeger inequality establish a fundamental connection between the Cheeger constant, which measures graph bottlenecks and global separability, and the second smallest eigenvalue of the graph Laplacian (also known as the Fiedler value or algebraic connectivity)\cite{chung1997spectral, fiedler1973algebraic}. In graph learning, such bottlenecks are important carriers of global structural information, as they determine how effectively signals can propagate across weakly connected regions and whether perturbations around bridge-like structures may alter the graph-level semantics\cite{topping2021understanding}. A larger Fiedler value generally indicates that the graph is more compact or less likely to be separated by a small number of edge cuts\cite{fiedler1973algebraic}. The Cheeger constant has been extensively studied in graph theory and has become an important metric for measuring graph structural stability and connectivity. However, to the best of our knowledge, there has been no research in GCL that utilizes the cheeger-inspired signature to define the graph's inherent invariance.

Beyond the standard graph Laplacian, the Hodge Laplacian extends spectral analysis to higher-order topological structures. While the graph Laplacian mainly describes node-level interactions and global connectivity, the Hodge Laplacian extends spectral analysis to edge-level and higher-order topological relations, making it particularly relevant to graph representation learning tasks involving nontrivial topological patterns. Early studies established the discrete geometric foundations of Hodge theory on graphs and complexes \cite{dimakis1994discrete, hiptmair2001discrete}, and subsequent works showed that Hodge Laplacians provide informative higher-order structural characterizations of networks \cite{johnson2012discrete, lim2020hodge}. Subsequent studies linked normalized Hodge Laplacians to higher-order random walks and diffusion processes, supporting their use in graph learning as effective descriptors of higher-order structure \cite{schaub2020random, wei2022hodge, wang2024heterophily, huang2025hl}. Compared with persistent-homology-based GCL methods such as TopoGCL and TensorMV-GCL\cite{wu2025tensor}, the Hodge Laplacian spectrum preserves operator-level spectral information over higher-order interactions and can therefore capture richer structural variation beyond persistence summaries alone. Meanwhile, both theoretical and empirical results indicate that the low-frequency Hodge spectrum contains informative signals about higher-order topological organization, including homological structure, higher-order connectivity, and community-level patterns \cite{aharoni2005eigenvalues, muravyev2025hodge, ribando2024combinatorial, krishnagopal2021spectral}. Motivated by these observations, we use low-frequency Hodge eigenvalues as intrinsic structural anchors for preserving higher-order topological invariance in graph contrastive learning.

\section{method}

\begin{figure*}
  \centering
  \includegraphics[width=\linewidth]{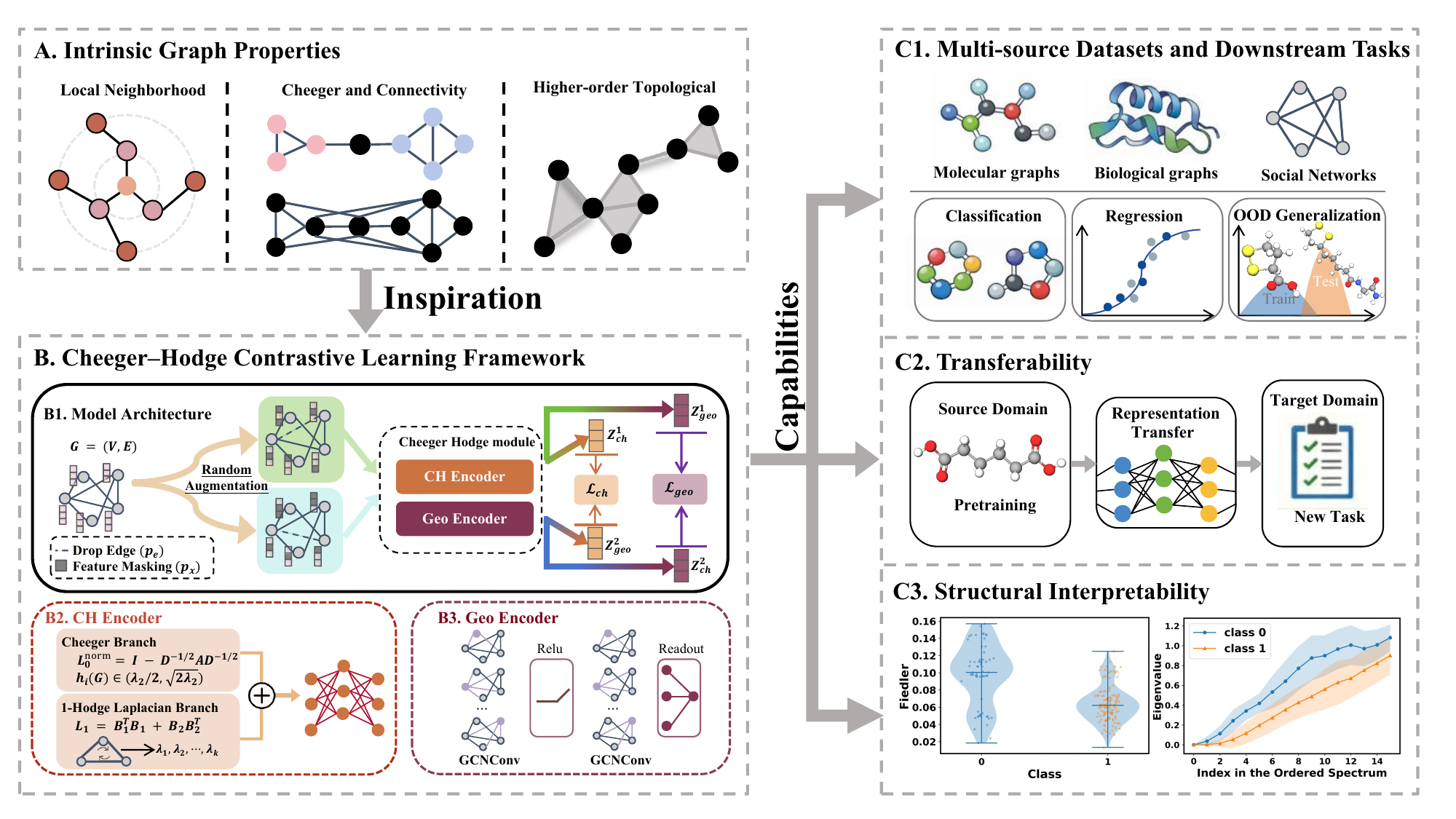}
  \caption{Schematic of CHCL and its capabilities on graph representation learning tasks. CHCL theory embeds Cheeger and Hodge structural priors into contrastive learning to reduce local bias, improve robustness to augmentation perturbations, and enhance structural awareness in learned representations. The framework combines a Cheeger--Hodge encoder with a geometric encoder, and jointly learns local geometric patterns and global structural properties through contrastive objectives. In this paper, CHCL is comprehensively evaluated across multiple datasets and downstream tasks, with additional analyses demonstrating its transferability and structural interpretability.}
  \label{fig:framework}
\end{figure*}

% In this section, we present the Cheeger--Hodge Contrastive Learning framework. We first analyze why Cheeger--Hodge structural signature is necessary for preserving graph-intrinsic invariance, and then describe how Cheeger--Hodge joint signature is incorporated into contrastive learning. The overall framework is illustrated in Figure~\ref{fig:framework} B.
In this section, we present the Cheeger--Hodge Contrastive Learning  framework. We first briefly recap graph contrastive learning, and then describe how the Cheeger--Hodge joint signature is incorporated into contrastive learning. The overall framework is illustrated in Figure~\ref{fig:framework} B.

\subsection{Preliminaries}

Let \(\mathcal{G}=(\mathcal{V},\mathcal{E},\mathbf{X})\) be an undirected graph, where \(\mathcal{V}\) is the node set, \(\mathcal{E}\) is the edge set, and \(\mathbf{X}\in\mathbb{R}^{|\mathcal{V}|\times F}\) denotes the node feature matrix with feature dimension \(F\). Let \(\mathbf{A}\in\mathbb{R}^{|\mathcal{V}|\times |\mathcal{V}|}\) be the adjacency matrix, where \(\mathbf{A}_{ij}=1\) if there is an edge between nodes \(i\) and \(j\), and \(\mathbf{A}_{ij}=0\) otherwise. Let \(\mathbf{D}\) be the corresponding degree matrix.

To construct augmented views for contrastive learning, we use two augmentation strategies: edge dropping and feature masking. For edge dropping, each edge is independently removed with probability \(p_e\), yielding an augmented adjacency matrix
\[
\mathbf{A}'=\mathbf{A}\odot \mathbf{R}_e,
\]
where \(\odot\) denotes the element-wise product and \(\mathbf{R}_e\) is a random binary mask matrix indicating whether an edge is retained. For feature masking, node features are independently masked with probability \(p_x\), yielding an augmented feature matrix
\[
\mathbf{X}'=\mathbf{X}\odot \mathbf{R}_x,
\]
where \(\mathbf{R}_x\) is a random binary mask matrix indicating whether a feature entry is retained.
% Applying these augmentations twice produces two augmented graph views, denoted by \(\mathcal{G}^{(1)}\) and \(\mathcal{G}^{(2)}\).

Consider the Laplacian matrix of the graph \( \mathcal{G} \), the symmetric normalized Laplacian is defined as
\begin{equation}
    L_0^{norm} = \mathbf{I} - \mathbf{D}^{-1/2} \mathbf{A} \mathbf{D}^{-1/2},
    \label{eq:L0}
\end{equation}
where  \( \mathbf{I} \) denotes the identity matrix. The spectrum of \(L_0^{norm} \) characterizes key global structural properties of the graph, such as connectivity and expansion, and serves as a fundamental operator in spectral graph analysis. 

To characterize higher-order structural information beyond node-level interactions, we consider the 1-Hodge Laplacian defined on edge space. Let \( \mathcal{G} \) be a graph with \( n = |\mathcal{V}| \) nodes and \( m = |\mathcal{E}| \) edges. The nodes are numbered arbitrarily but fixed as \( \{1, \dots, n\} \), and we denote the oriented edge set by \(\{(u_e,v_e)\}_{e=1}^m\). Based on this orientation, the node–edge incidence matrix \( B_1 \in \mathbb{R}^{n \times m} \) is defined as
\begin{equation}
(B_1)_{i,e}
=
\begin{cases}
-1, & \text{if } i=u_e,\\
+1, & \text{if } i=v_e,\\
0, & \text{otherwise}.
\end{cases}
\end{equation}

Next, we enumerate all triangles \( (a, b, c) \) in the edge set and denote the resulting triangle set by \(\{\triangle_t\}_{t=1}^{\mathcal{T}}\), where $\mathcal{T}$ is the total number of triangles. For each triangle, as shown in the 1-Hodge Laplacian Branch in Figure \ref{fig:framework} B2, we independently choose a cycle direction and construct the edge-triangle incidence matrix \( B_2 \in \mathbb{R}^{m \times \mathcal{T}} \) as:
\begin{equation}
(B_2)_{e,t}
=
\begin{cases}
+1, & e\in \partial \triangle_t \text{ and } \operatorname{ori}(e)=\operatorname{ori}_t(e),\\
-1, & e\in \partial \triangle_t \text{ and } \operatorname{ori}(e)\neq \operatorname{ori}_t(e),\\
0, & e\notin \partial \triangle_t.
\end{cases}
\end{equation}
where \(\partial \triangle_t\) denotes the set of boundary edges of triangle \(\triangle_t\), and \(\operatorname{ori}_t(e)\) denotes the orientation induced by \(\triangle_t\) on edge \(e\).

The 1-Hodge Laplacian is then defined as:
\begin{equation}
L_1 = B_1^\top B_1 + B_2 B_2^\top \in \mathbb{R}^{m \times m},
\label{eq:L1}
\end{equation}
which operates on the edge space and has an orthogonal decomposition into gradient, curl, and harmonic subspaces. Accordingly, the spectrum of \(L_1\), especially its low-frequency part, characterizes higher-order structural information of the graph through edge-flow patterns associated with these subspaces.

\subsection{Cheeger--Hodge Contrastive Learning}
Given a dataset \( \mathcal{D} = \{\mathcal{G}_1, \mathcal{G}_2, \dots, \mathcal{G}_N\} \), where each graph is denoted by \(\mathcal{G}_i=(\mathcal{V}_i,\mathcal{E}_i,X_i)\), we generate two augmented views for each graph using edge dropping and feature masking, denoted by
$
\mathcal{G}_i^{(1)}=(\mathcal{V}_i,\mathcal{E}_i^{(1)},\mathbf{X}_i^{(1)}) \quad \text{and} \quad
\mathcal{G}_i^{(2)}=(\mathcal{V}_i,\mathcal{E}_i^{(2)},\mathbf{X}_i^{(2)}).
$
% we generate two augmented views for each graph, denoted by \(\mathcal{G}_i^{(1)}\) and \(\mathcal{G}_i^{(2)}\), using edge dropping and feature masking. 

\subsubsection{Geometry branch}
As illustrated in Fig.~\ref{fig:framework} B3, each augmented view is encoded by a Graph Convolutional Network (GCN) to obtain graph-level embeddings. For the two augmented views \(\mathcal{G}_i^{(1)}\) and \(\mathcal{G}_i^{(2)}\), the node representations are updated through \(L\) layers of message passing:
\begin{equation}
\begin{split}
    \mathbf{H}_i^{(\ell+1,1)}
&=
\mathrm{MP}^{(\ell)}\!\left(\mathbf{H}_i^{(\ell,1)},\mathcal{G}_i^{(1)}\right),\\
\mathbf{H}_i^{(\ell+1,2)}
&=
\mathrm{MP}^{(\ell)}\!\left(\mathbf{H}_i^{(\ell,2)},\mathcal{G}_i^{(2)}\right),
\end{split}
\end{equation}
for \(\ell=0,\ldots,L-1\), where
\[
\mathbf{H}_i^{(0,1)}=\mathbf{X}_i^{(1)},
\qquad
\mathbf{H}_i^{(0,2)}=\mathbf{X}_i^{(2)}.
\]
The final graph-level embeddings are then obtained through a readout function:
\begin{equation}
\begin{split}
    z_{\mathrm{geo},i}^{(1)}
&=
\mathrm{READOUT}\!\left(\mathbf{H}_i^{(L,1)}\right),\\
z_{\mathrm{geo},i}^{(2)}
&=
\mathrm{READOUT}\!\left(\mathbf{H}_i^{(L,2)}\right).
\end{split}
\end{equation}

\subsubsection{Cheeger branch}
The Cheeger branch is built on the spectrum of the normalized graph Laplacian \(L_0^{norm}(\mathcal G)\) in Eq.~\eqref{eq:L0}. Denote its eigenvalues by
\begin{equation}
0=\lambda_1\leq \lambda_2\leq \cdots \leq \lambda_n.
\end{equation}

By the classical Cheeger inequality, \(\lambda_2\) provides upper and lower bounds for the Cheeger constant \(h(\mathcal G)\), thereby relating the graph's expansion properties and bottleneck structure to the Fiedler value. When \( \lambda_2 \) is large, the graph is harder to partition into disconnected subgraphs, while when \( \lambda_2 \) is close to zero, the graph exhibits bottleneck or sparse cuts. Therefore, \(\lambda_2\) serves as a stable and computable geometric proxy for graph expansion without explicitly searching for all cuts. Compared to relying solely on statistics like node degrees, \( \lambda_2 \) integrates global connectivity patterns through spectral decomposition, making it a more informative geometric indicator for capturing graph expansion and clustering structure.

According to the Cheeger inequality, the Cheeger constant \(h(\mathcal G)\) satisfies
\begin{equation}\label{eq:cheeger}
\frac{\lambda_2}{2}
\le
h(\mathcal G)
\le
\sqrt{2\,\lambda_2}.
\end{equation}

Motivated by this interval characterization, we define a deterministic Cheeger signature vector induced by \(\lambda_2\) as
\begin{equation}
\mathbf h_c(\mathcal G)
% =
% \phi\!\left(\lambda_2\right)
=
\bigl[h_{c,1}(\mathcal G),\dots,h_{c,d_c}(\mathcal G)\bigr]^\top,
\end{equation}
where each component is given by
\begin{equation}
% \begin{aligned}
h_{c,j}(\mathcal G)=
% &=
\frac{\lambda_2}{2} + 
% &\quad+
\frac{j-1}{d_c-1}
\left(
\sqrt{2\,\lambda_2}
-
\frac{\lambda_2}{2}
\right),
% \end{aligned}
\end{equation}
where \(j=1,\dots,d_c\).

Rather than using the scalar \(\lambda_2\) directly, we map it to a \(d_c\)-dimensional signature vector for two reasons. First, a higher-dimensional encoding is more suitable for joint fusion with the Hodge signature. Second, the interval \([\lambda_2/2,\sqrt{2\lambda_2}]\) is naturally induced by the Cheeger inequality and therefore provides a deterministic structural range associated with the graph's connectivity profile. Uniformly sampling this interval yields a simple and stable signature representation that preserves Cheeger-related information without requiring explicit computation of the Cheeger constant.

\subsubsection{1-Hodge Laplacian branch}
The 1-Hodge Laplacian Branch leverages the low-frequency spectrum of the 1-Hodge Laplacian defined in Eq.~\eqref{eq:L1} to characterize higher-order topological properties of the graph. To this end, we perform sparse eigendecomposition on $L_1$ and retain the smallest $d_h$ non-negative eigenvalues:
\begin{equation}
0 \le \mu_1 \le \mu_2 \le \cdots \le \mu_{d_h}.
\end{equation}

To avoid numerical instability with small eigenvalues, we apply a \( \log(1+\cdot) \) transformation to these eigenvalues and obtain the Hodge signature:
\begin{equation}\label{eq:hodge}
\mathbf{h}_h(\mathcal{G})=\big[\log(1+\mu_1),\log(1+\mu_2),\ldots,\log(1+\mu_{d_h})\big].
\end{equation}

This transformation preserves the relative differences among small eigenvalues while compressing the scale of larger ones.

\subsubsection{Cheeger--Hodge joint signature}
We then concatenate the Cheeger signature \( \mathbf{h}_c(\mathcal{G}) \) and the Hodge signature  \( \mathbf{h}_h(\mathcal{G}) \) to form the Cheeger--Hodge joint signature:
\begin{equation}
\mathbf{h}_{\text{CH}}(\mathcal{G})=\mathbf{h}_c(\mathcal{G})\,\|\,\mathbf{h}_h(\mathcal{G}).
\end{equation}

For the two augmented views \(\mathcal{G}_i^{(1)}\) and \(\mathcal{G}_i^{(2)}\), the corresponding Cheeger--Hodge joint signatures \(\mathbf{h}_{\mathrm{CH},i}^{(1)}\) and \(\mathbf{h}_{\mathrm{CH},i}^{(2)}\) are passed through an MLP projection head to obtain latent embeddings
\begin{equation}
z_{\mathrm{CH},i}^{(1)}=\mathrm{MLP}\!\left(\mathbf{h}_{\mathrm{CH},i}^{(1)}\right), 
\qquad
z_{\mathrm{CH},i}^{(2)}=\mathrm{MLP}\!\left(\mathbf{h}_{\mathrm{CH},i}^{(2)}\right).
\end{equation}

In this way, the Cheeger--Hodge branch provides a perturbation-stable structural representation that complements the encoder-learned geometric representation.

\subsubsection{Training objective}
We employ the NT-Xent loss function, a widely used contrastive learning objective for measuring the similarity between graph embeddings. For each anchor embedding \(z_{\mathrm{geo},i}^{(1)}\), the embedding \(z_{\mathrm{geo},i}^{(2)}\) from the other view of the same graph is treated as the positive sample, while \(\{z_{\mathrm{geo},j}^{(2)}\}_{j\neq i}\) are treated as negative samples. The resulting geometry-level contrastive loss is
\begin{equation}
\mathcal{L}_{\mathrm{geo},i}
=
-
\log
\frac{
\exp\!\bigl(\mathrm{sim}(z_{\mathrm{geo},i}^{(1)},z_{\mathrm{geo},i}^{(2)})/\tau\bigr)
}{
\sum_{j=1,\;j\neq i}^{N}
\exp\!\bigl(\mathrm{sim}(z_{\mathrm{geo},i}^{(1)},z_{\mathrm{geo},j}^{(2)})/\tau\bigr)
},
\end{equation}
where \(\mathrm{sim}(\cdot,\cdot)\) denotes cosine similarity and \(\tau\) is a temperature hyperparameter.

Similarly, we apply the same contrastive objective to the Cheeger--Hodge embeddings:
\begin{equation}
\mathcal{L}_{\mathrm{CH},i}
=
-
\log
\frac{
\exp\!\bigl(\mathrm{sim}(z_{\mathrm{CH},i}^{(1)},z_{\mathrm{CH},i}^{(2)})/\tau\bigr)
}{
\sum_{j=1,\;j\neq i}^{N}
\exp\!\bigl(\mathrm{sim}(z_{\mathrm{CH},i}^{(1)},z_{\mathrm{CH},j}^{(2)})/\tau\bigr)
}.
\end{equation}

The total objective is then given by
\begin{equation}
\mathcal{L}
=
\lambda_{\mathrm{geo}} \frac{1}{N}\sum_{i=1}^{N}\mathcal{L}_{\mathrm{geo},i}
+
\lambda_{\mathrm{CH}} \frac{1}{N}\sum_{i=1}^{N}\mathcal{L}_{\mathrm{CH},i},
\label{eq:loss}
\end{equation}
where \( \lambda_{\text{geo}} \) and \( \lambda_{\text{CH}} \) are trade-off hyperparameters. By minimizing Eq.~\eqref{eq:loss}, CHCL jointly aligns the local geometric representations learned by the GCN with the perturbation-stable structural representations produced by the Cheeger--Hodge branch, yielding graph-level representations that preserve both geometric and topological properties.

\section{Theoretical analysis}

In this section, we provide the theoretical analysis of CHCL by formalizing the Cheeger--Hodge joint signature and studying its structural properties. 
% We begin by introducing the definition of the Cheeger--Hodge joint signature, and then analyze its complementarity and stability.

\begin{definition}\label{def:cheeger_hodge_signature}
Let \( \mathcal{G} = (\mathcal{V}, \mathcal{E}) \) be a simple graph, and \( \mathcal{K} \) be a 2-dimensional simplicial complex on the vertex set \( \mathcal{V} \). Let \( L_0^{norm} \) denote the normalized graph Laplacian of \( \mathcal{G} \), and \( L_1 \) denote the 1-Hodge Laplacian of \( \mathcal{K} \). The eigenvalues of \( L_0^{norm} \) and \( L_1 \) are ordered in non-decreasing order as:
\[
0 = \lambda_1(L_0^{norm}) \le \lambda_2(L_0^{norm}) \le \cdots \le \lambda_n(L_0^{norm}) , 
\] 
\[
0 \le \mu_1(L_1) \le \mu_2(L_1) \le \cdots \le \mu_m(L_1).
\]  
Let \(\phi:[\delta,\infty)\to\mathbb{R}^{d_c}\) denote the deterministic Cheeger signature map induced by \(\lambda_2\). Let \(d_h\) denote the number of retained low-frequency Hodge eigenvalues, with \(1\le d_h\le m\). Define
\[
\mathbf{h}_c(\mathcal{G})
:=
\phi\!\left(\lambda_2\bigl(L_0^{norm}\bigr)\right)
\in\mathbb{R}^{d_c},
\]
and
\[
\mathbf{h}_{\mu}(\mathcal{G})
:=
\bigl[
\log\bigl(1+\mu_1(L_1)\bigr),
\ldots,
\log\bigl(1+\mu_{d_h}(L_1)\bigr)
\bigr]^\top
\in\mathbb{R}^{d_h}.
\]

We then define the Cheeger--Hodge joint signature as
\[
\mathbf{h}_{\mathrm{CH}}(\mathcal{G})
:=
\begin{bmatrix}
\mathbf{h}_c(\mathcal{G})\\[1mm]
\mathbf{h}_{\mu}(\mathcal{G})
\end{bmatrix}
\in\mathbb{R}^{d_c+d_h}.
\]
\end{definition}

% [Extended Complementarity of Cheeger and Hodge Signatures]
% \begin{theorem}
% Let \(\mathbf{h}_c(\mathcal{G})\) and \(\mathbf{h}_{\mu}(\mathcal{G})\) be defined as in Definition~\ref{def:cheeger_hodge_signature}. Under the Hodge decomposition of the edge space,
% \[
% \mathbb{R}^{|\mathcal{E}|}
% =
% \operatorname{im}(B_1^\top)\oplus \ker(L_1)\oplus \operatorname{im}(B_2),
% \]
% the retained low-frequency Hodge modes decompose along three orthogonal components: the gradient component \(\operatorname{im}(B_1^\top)\), the harmonic component \(\ker(L_1)=\ker(B_1)\cap\ker(B_2^\top)\), and the curl component \(\operatorname{im}(B_2)\). Moreover, the spectrum of \(L_1\) on \(\operatorname{im}(B_1^\top)\) coincides exactly with the nonzero spectrum of \(L_0\), so the Hodge signature \(\mathbf{h}_{\mu}(\mathcal{G})\) should be understood as a structured extension of the Cheeger signature \(\mathbf{h}_c(\mathcal{G})\). 
% % Any potential redundancy is confined to the gradient branch, while the harmonic and curl branches contribute genuinely new low-frequency structural information that cannot be captured by the Cheeger descriptor alone.
% In particular, if
% \[
% \mu_{d_h}(L_1)<\lambda_2(L_0),
% \]
% then none of the retained Hodge modes belongs to the gradient branch, and hence \(\mathbf{h}_c(\mathcal{G})\) and \(\mathbf{h}_{\mu}(\mathcal{G})\) are strictly complementary.
% \end{theorem}

\section{EXPERIMENTS}

In this section, we conduct a comprehensive evaluation of CHCL across diverse graph-level learning settings. These experiments examine the quality of the learned representations from multiple perspectives, including their effectiveness in graph-level prediction, and their transferability. We also conduct in-depth analyses of stability, robustness and interpretability to gain a better understanding of CHCL's behaviour and reliability.

\subsection{Dataset and Baselines}
\subsubsection{Datasets}
We evaluate CHCL under multiple learning settings using a diverse collection of benchmark graph datasets. 
For both unsupervised and semi-supervised learning, we use the following 10 real-world graph datasets from the TUdataset \cite{Morris+2020}. 
For unsupervised graph regression, we adopt three molecular property prediction datasets\cite{hu2020open}.
For transfer learning, we pretrain the model on the large-scale ChEMBL\cite{gaulton2012chembl} dataset and fine-tune it on eight downstream MoleculeNet benchmarks\cite{wu2018moleculenet}.
Evaluation follows 10-fold cross-validation on the TUdataset benchmarks and the official data splits on all other datasets. 

\subsubsection{Baselines}
To evaluate the performance of the proposed CHCL model, we compare it with a diverse set of 21 baseline methods, covering various categories of graph representation learning techniques. These include vanilla, adaptive, and structure-aware graph contrastive learning methods for graph-level prediction; graph pretraining methods for transfer learning. The reported results are directly obtained from the baseline methods' published papers. For results not available in the original publications, we conducted experiments based on their open-source code.

\subsection{Unsupervised Learning}
\begin{table*}[htbp]
\centering
\caption{Performance comparison on benchmark graph classification datasets. Results are reported as mean $\pm$ standard deviation.\textbf{Bold} denotes the best performance, and \underline{underline} represents the second best performance. * indicates results that we reproduced.
}
\label{tab:result}
\resizebox{\textwidth}{!}{
\begin{tabular}{lccccccccccc}
\toprule
\textbf{Model} & \textbf{PROTEINS} & \textbf{DD} & \textbf{MUTAG} & \textbf{BZR} & \textbf{COX2} & \textbf{PTC\_MR} & \textbf{PTC\_FM} & \textbf{IMDB-B} & \textbf{IMDB-M} & \textbf{REDDIT-B} \\
\midrule
% GL          & N/A & N/A & 81.66$\pm$2.11 & N/A & N/A & 57.30$\pm$1.40 & N/A & 65.87$\pm$0.98 & 46.50$\pm$0.30 & 77.34$\pm$0.18\\
% WL          & 72.92$\pm$0.56 & 74.00$\pm$2.20 & 80.72$\pm$3.00 & N/A & N/A & 58.00$\pm$0.50 & N/A & 72.30$\pm$3.44 & 47.00$\pm$0.50 & 68.82$\pm$0.41\\
% DGK         & 73.30$\pm$0.82 & N/A & 87.44$\pm$2.72 & N/A & N/A & 60.10$\pm$2.60 & N/A & 66.96$\pm$0.56 & 44.60$\pm$0.50 & 78.04$\pm$0.39\\
% node2vec    & 57.49$\pm$3.57 & N/A & 72.63$\pm$10.20 & N/A & N/A & N/A & N/A & 56.40$\pm$2.80 & 36.00$\pm$0.70 & 69.70$\pm$4.10\\
% sub2vec     & 53.03$\pm$5.55 & N/A & 61.05$\pm$15.80 & N/A & N/A & N/A & N/A & 55.26$\pm$1.54 & 36.70$\pm$0.80 & 71.48$\pm$0.41\\
% graph2vec   & 73.30$\pm$2.05 & N/A & 83.15$\pm$9.25 & N/A & N/A & N/A & N/A & 71.10$\pm$0.54 & 50.40$\pm$0.90 & 75.78$\pm$1.03\\
InfoGraph   & 74.44$\pm$0.31 & 72.85$\pm$1.78 & 89.01$\pm$1.13 &  84.84$\pm$0.86 & 80.55$\pm$0.51 & 61.70$\pm$1.40 & 61.55$\pm$0.92 & 73.03$\pm$0.87 & 49.70$\pm$0.50 &82.50$\pm$1.42\\
GraphCL     & 74.39$\pm$0.45 & 78.62$\pm$0.40 & 86.80$\pm$1.34 &  84.20$\pm$0.86 & 81.10$\pm$0.82 & 61.30$\pm$2.10 & 65.26$\pm$0.59 & 71.14$\pm$0.44 & 49.20$\pm$0.60 &89.53$\pm$0.84\\
AD-GCL      & 73.28$\pm$0.46 & 75.79$\pm$0.87 & 88.74$\pm$1.85 &  85.97$\pm$0.63 & 78.68$\pm$0.56 & 63.20$\pm$1.76 & 64.99$\pm$0.77 & 70.21$\pm$0.68 & 50.60$\pm$0.70 &90.07$\pm$0.85\\
AutoGCL     & 75.80$\pm$0.36 & 77.57$\pm$0.60 & 88.64$\pm$1.08 &  86.27$\pm$0.71 & 79.31$\pm$0.70 & 63.10$\pm$2.30 & 63.62$\pm$0.55 & 72.32$\pm$0.93 & 50.60$\pm$0.80 &88.58$\pm$1.49\\
RGCL        & 73.03$\pm$0.43 & 78.86$\pm$0.48 & 87.66$\pm$1.01 &  84.54$\pm$1.67 & 79.31$\pm$0.68 & 61.43$\pm$0.52 & 64.29$\pm$0.32 & 71.85$\pm$0.84 & 49.31$\pm$0.42 &90.34$\pm$0.58\\
JOAO        & 74.55$\pm$0.41 & 75.69$\pm$0.67 & 87.35$\pm$1.02 &  83.02$\pm$1.28* & 78.42$\pm$1.65* & 57.17$\pm$2.12* & 59.98$\pm$2.17* & 70.21$\pm$3.08 & 47.03$\pm$2.29* &85.29$\pm$1.35\\
% SEGA        & 76.01$\pm$0.42 & 78.76$\pm$0.57 & 90.21$\pm$0.66 &   &  &  &  & 73.58$\pm$0.44 &  &90.21$\pm$0.65\\
GCL-SPAN & 75.78$\pm$0.41 & 75.78$\pm$0.52 & 89.12$\pm$0.76 & \underline{87.80$\pm$1.54*} & 82.08$\pm$1.67* & 62.86$\pm$4.04* & 66.67$\pm$3.04* & 73.65$\pm$0.69 & 52.16$\pm$0.72 &83.62$\pm$0.64\\
TopoGCL & \underline{77.30$\pm$0.89} & 79.15$\pm$0.35 & 90.09$\pm$0.93 &  87.17$\pm$0.83 & 81.45$\pm$0.55 & \underline{63.43$\pm$1.13} & 67.11$\pm$1.08 & 74.67$\pm$0.32 & \underline{52.81$\pm$0.31} &90.40$\pm$0.53\\
CI-GCL & 76.50$\pm$0.10 & \underline{79.63$\pm$0.30} & 89.67$\pm$0.90 & 86.70$\pm$1.01* & \underline{84.11$\pm$1.45*} & 61.60$\pm$1.60* & 64.53$\pm$1.27* & 73.85$\pm$0.80 & 50.21$\pm$0.69* &90.80$\pm$0.5\\
% SAGCL & 76.88$\pm$0.24 & N/A & \underline{91.82$\pm$0.57} & N/A & N/A & \underline{63.83$\pm$0.76} & N/A & 74.08$\pm$0.63 & 51.73$\pm$0.44 &90.65$\pm$0.21\\

GMCL & 77.16$\pm$0.73 & 78.68$\pm$0.28 & \textbf{93.58$\pm$2.13} & 84.44$\pm$5.17* & 79.02$\pm$1.47* & 61.65$\pm$4.69* & \underline{68.47$\pm$8.02*} & \underline{75.48$\pm$0.52} & 49.93$\pm$2.71* &\underline{92.39$\pm$0.65}\\
\midrule
CH-CL & \textbf{79.06$\pm$0.58} & \textbf{80.92$\pm$0.85} & \underline{93.02$\pm$0.61} & \textbf{89.19$\pm$0.71} & \textbf{84.20$\pm$0.76} & \textbf{68.18$\pm$0.56} & \textbf{68.90$\pm$1.07} & \textbf{75.62$\pm$0.61} & \textbf{53.20$\pm$0.53}&\textbf{92.49$\pm$0.43} \\
\bottomrule
\end{tabular}}
\end{table*}

\begin{table}[ht]
\centering
\caption{RMSE for unsupervised graph regression. Results are reported as mean $\pm$ standard deviation. \textbf{Bold} denotes the best performance, and \underline{underline} represents the second best performance.}
\label{tab:regression}
\begin{tabular}{lccc c}
\toprule
\textbf{Method} & \textbf{molesol} & \textbf{mollipo} & \textbf{molfreesolv}  \\
\midrule
InfoGraph      & 1.34$\pm$0.18 & 1.01$\pm$0.02 & 10.0$\pm$4.82  \\
GraphCL        & 1.27$\pm$0.09 & 0.91$\pm$0.02 & 7.68$\pm$2.75  \\
% MVGRL          & 1.43$\pm$0.15 & 0.96$\pm$0.04 & 9.02$\pm$1.98  \\
JOAO           & 1.29$\pm$0.12 & 0.87$\pm$0.03 & 5.13$\pm$0.72  \\
GCL-SPAN       & 1.22$\pm$0.05 & \underline{0.80$\pm$0.02} & 4.53$\pm$0.46  \\
AD-GCL         & 1.22$\pm$0.09 & 0.84$\pm$0.03 & 5.15$\pm$0.62  \\
CI-GCL         & \underline{1.13$\pm$0.13} & 0.82$\pm$0.03 & \underline{2.87$\pm$0.32}  \\
\midrule
CHCL & \textbf{0.93$\pm$0.02} & \textbf{0.78$\pm$0.01} & \textbf{2.59$\pm$0.12}  \\
\bottomrule
\end{tabular}

\end{table}

To comprehensively evaluate the effectiveness of our model, we follow the experimental setup of \cite{tan2024community}. For all graph datasets, we use an SVM as the evaluation model and report the resulting performance. All experiments are repeated five times, and we report the mean and standard deviation. We use the TU datasets \cite{Morris+2020} and OGB datasets \cite{hu2020open} to evaluate graph classification and regression tasks, respectively. Table \ref{tab:result} presents the performance on graph classification, while Table \ref{tab:regression} shows the performance on graph regression. These results highlight the effectiveness of the proposed model across multiple domains.

As shown in Table \ref{tab:result}, CHCL achieves the best performance on 9 out of 10 graph classification datasets and ranks second on the remaining MUTAG dataset. Its consistently competitive results across chemical compounds, molecular compounds, and social networks demonstrate that CHCL learns a generalizable and discriminative representation. As shown in Table \ref{tab:regression}, CHCL also attains the lowest RMSE on three unsupervised molecular graph regression datasets, demonstrating strong cross-task generalization. On semantically sensitive and structurally decisive datasets such as PTC\_MR and PROTEINS, CHCL yields substantial improvements, outperforming the strongest baseline by 7.5\% and 2.3\%, respectively. Meanwhile, on datasets where existing methods already perform strongly, CHCL still delivers superior performance. These results suggest that incorporating structural alignment helps improve representation quality. Moreover, CHCL consistently ranks first on all three unsupervised graph regression datasets, with particularly large RMSE reductions on molesol and molfreesolv, demonstrating that the learned representations capture continuous structure–property relationships rather than being merely effective for classification.

Notably, the four strongest baselines are GCL-SPAN, TopoGCL, CI-GCL, and GMCL, among which GCL-SPAN, TopoGCL, and CI-GCL are structure- or topology-aware GCL methods. Overall, methods that explicitly incorporate structural and topological information obtain the best performance on more datasets, highlighting the importance of modeling structure and topology for improving graph representation learning. 
% Compared with the state-of-the-art graph contrastive learning baselines GCL-SPAN, TopoGCL, CI-GCL, and GMCL, CHCL achieves average improvements of approximately 4.17\%, 2.78\%, 3.59\%, and 3.15\%, respectively.

% These results reflect the effectiveness of the Cheeger--Hodge signatures in helping the model capture the intrinsic properties of the graph. Specifically, on the PROTEINS dataset, CH-CL achieves 78.99$\pm$0.58, surpassing the previous best result of 77.30 from TopoGCL by +1.69. On MUTAG, CH-CL reaches 93.02$\pm$0.61, outperforming the prior state-of-the-art of 91.82 from SAGCL by +1.20. On the COX2 dataset, CH-CL achieves 84.20$\pm$0.76, surpassing the previous best result of 81.45 from TopoGCL by +2.75. Among the five baseline categories, the strongest baseline are all structure-aware or topology-aware GCL methods. TopoGCL achieves the best results in six datasets, SAGCL in two, and CI-GCL in one. This highlights the crucial role of explicitly modeling structure and topology in enhancing performance. Compared with the state-of-the-art graph contrastive learning methods TopoGCL, CI-GCL, and SAGCL, the CH-CL method shows average improvements of approximately 1.37\%, 2.10\%, 1.18\%, respectively. 
% These results underscore the potential of integrating Cheeger--Hodge signatures into contrastive learning frameworks, providing a powerful tool for capturing and leveraging the intrinsic properties of graphs.

\subsection{Transfer Learning}
To evaluate transferability, we conduct unsupervised pretraining on the large-scale molecular graph dataset ChEMBL. We then use the pretrained parameters to initialize the model and perform supervised fine-tuning on eight MoleculeNet molecular property prediction benchmarks. During fine-tuning, a task-specific prediction head is attached to the pretrained model and the entire network is optimized end-to-end. Table \ref{tab:transfer} reports the mean $\pm$ standard deviation of ROC-AUC over five repeated runs.

As shown in Table \ref{tab:transfer}, pretraining consistently improves downstream performance, indicating that the structural knowledge acquired during pretraining can be effectively carried over to supervised fine-tuning. Compared with existing pretraining methods, CHCL achieves the strongest overall performance on most datasets and remains competitive on the rest. These results show that incorporating global structure and higher-order topology during pretraining leads to more transferable representations, which provide a better foundation for downstream adaptation across diverse molecular property prediction tasks. Moreover, the stable performance of CHCL across different benchmarks further highlights its strong transferability across diverse downstream data distributions.
% As shown in Table \ref{tab:transfer}, pretraining substantially improves downstream performance. Compared with No Pretrain, CHCL improves ROC-AUC on all eight datasets with an average relative gain of 12.40\%. The improvement is particularly pronounced on ClinTox, where CHCL achieves a relative gain of 44.10\%. This observation suggests that the structural invariances captured during pretraining are especially beneficial for tasks involving semantically complex molecular distinctions.
% Against existing pretraining methods, CHCL achieves the best results on 5 out of 8 datasets and ranks second on HIV. Notably, CHCL surpasses the strongest competing method by 1.58\% on SIDER and 2.61\% on BACE, suggesting that explicitly modeling global structure and higher-order topology during pretraining is particularly beneficial for tasks that depend on global molecular organization and topological motifs. While CHCL is not the top performer on MUV and Tox21, it remains competitive and exhibits relatively small variance, supporting its robustness across diverse downstream distributions.

\begin{table*}[ht]
\centering
\caption{Comparison with the existing methods for transfer learning.  Results are reported as mean $\pm$ standard deviation. \textbf{Bold} denotes the best performance, and \underline{underline} represents the second best performance. * indicates results that we reproduced.}
\label{tab:transfer}
\begin{tabular}{lcccccccc}
\toprule
\textbf{Model} & \textbf{BBBP} & \textbf{Tox21} & \textbf{ToxCast} & \textbf{SIDER} & \textbf{ClinTox} & \textbf{MUV} & \textbf{HIV} & \textbf{BACE} \\
\midrule
No Pretrain     & 65.8$\pm$4.5  & 74.0$\pm$0.8  & 63.4$\pm$0.6  & 57.3$\pm$1.6  & 58.0$\pm$4.4  & 71.8$\pm$2.5  & 75.3$\pm$1.9  & 70.1$\pm$5.4 \\
\midrule
Infomax        & 68.8$\pm$0.8  & 75.3$\pm$0.5  & 62.7$\pm$0.4  & 58.4$\pm$0.8  & 69.9$\pm$3.0  & 75.3$\pm$2.5  & 76.0$\pm$0.7  & 75.9$\pm$1.6 \\
EdgePred       & 67.3$\pm$2.4  & 76.0$\pm$0.6  & 64.1$\pm$0.6  & 60.4$\pm$0.7  & 64.1$\pm$3.7  & 74.1$\pm$2.1  & 76.3$\pm$1.0  & 79.9$\pm$0.9 \\
AttrMasking    & 64.3$\pm$2.8  & \textbf{76.7$\pm$0.4}  & 64.2$\pm$0.5  & 61.0$\pm$0.7  & 71.8$\pm$4.4  & 74.7$\pm$1.4  & 77.2$\pm$1.1  & 79.3$\pm$1.6 \\
ContextPred    & 68.0$\pm$2.0  & 75.7$\pm$0.7  & 63.9$\pm$0.6  & 60.9$\pm$0.6  & 65.9$\pm$3.8  & 75.8$\pm$1.7  & 77.3$\pm$1.0  & 79.6$\pm$1.2 \\
GraphCL        & 69.68$\pm$0.67 & 73.87$\pm$0.66 & 62.40$\pm$0.57 & 60.53$\pm$0.88  & 75.99$\pm$2.65  & 69.80$\pm$2.66  & 78.47$\pm$1.22  & 75.38$\pm$1.44 \\
JOAO         & 70.22$\pm$0.98 & 74.98$\pm$0.29 & 62.94$\pm$0.48 & 59.97$\pm$0.79  & 81.32$\pm$2.49  & 71.66$\pm$1.43  & 76.73$\pm$1.23  & 77.34$\pm$0.48 \\
% JOAOv2         & 71.39$\pm$0.92 & 74.27$\pm$0.62 & 63.16$\pm$0.45 & 60.49$\pm$0.74  & 80.97$\pm$1.64  & 73.67$\pm$1.00  & 77.51$\pm$1.17  & 75.49$\pm$1.27 \\
AD-GCL         & 70.01$\pm$1.07 & \underline{76.54$\pm$0.82} & 63.07$\pm$0.72 & \underline{63.28$\pm$0.79} & 79.78$\pm$3.52 & 72.30$\pm$1.61  & 78.28$\pm$0.97  & 78.51$\pm$0.80 \\
AutoGCL           & 73.36$\pm$0.77 & 75.69$\pm$0.29 & 63.47$\pm$0.38 & 62.51$\pm$0.63 & 80.99$\pm$3.38 & \underline{75.83$\pm$1.30} & 78.35$\pm$0.64 & \underline{83.26$\pm$1.13} \\
GMCL          & \underline{73.80$\pm$1.21} & 74.94$\pm$0.85* & \underline{64.21$\pm$0.33} & 61.75$\pm$1.02* & \underline{83.37$\pm$3.83} & \textbf{78.03$\pm$1.56} & \textbf{79.88$\pm$0.76} & 82.37$\pm$1.16 \\
\midrule
CHCL           & \textbf{74.33$\pm$0.62} & 76.06$\pm$0.22 & \textbf{64.54$\pm$0.16} & \textbf{64.28$\pm$0.35} & \textbf{83.58$\pm$1.53} & 75.43$\pm$0.85 & \underline{78.50$\pm$0.46} & \textbf{85.43$\pm$0.62} \\
\bottomrule
\end{tabular}

\end{table*}

\subsection{Ablation}
\subsubsection{Analysis of Module Contributions}
To verify the impact of each component on the model performance, we conduct ablation studies to investigate the contributions of different modules in our framework. Specifically, we consider three ablation settings: (1) \textbf{w/o Cheeger:} removing the Cheeger module, (2) \textbf{w/o Hodge:} removing the Hodge module, and (3) \textbf{w/o CH:} removing both the Cheeger and Hodge modules simultaneously. These ablations allow us to quantify how each component affects the final performance and thus provide deeper insights into their roles. 

\begin{figure}
  \centering
  \includegraphics[width=\linewidth]{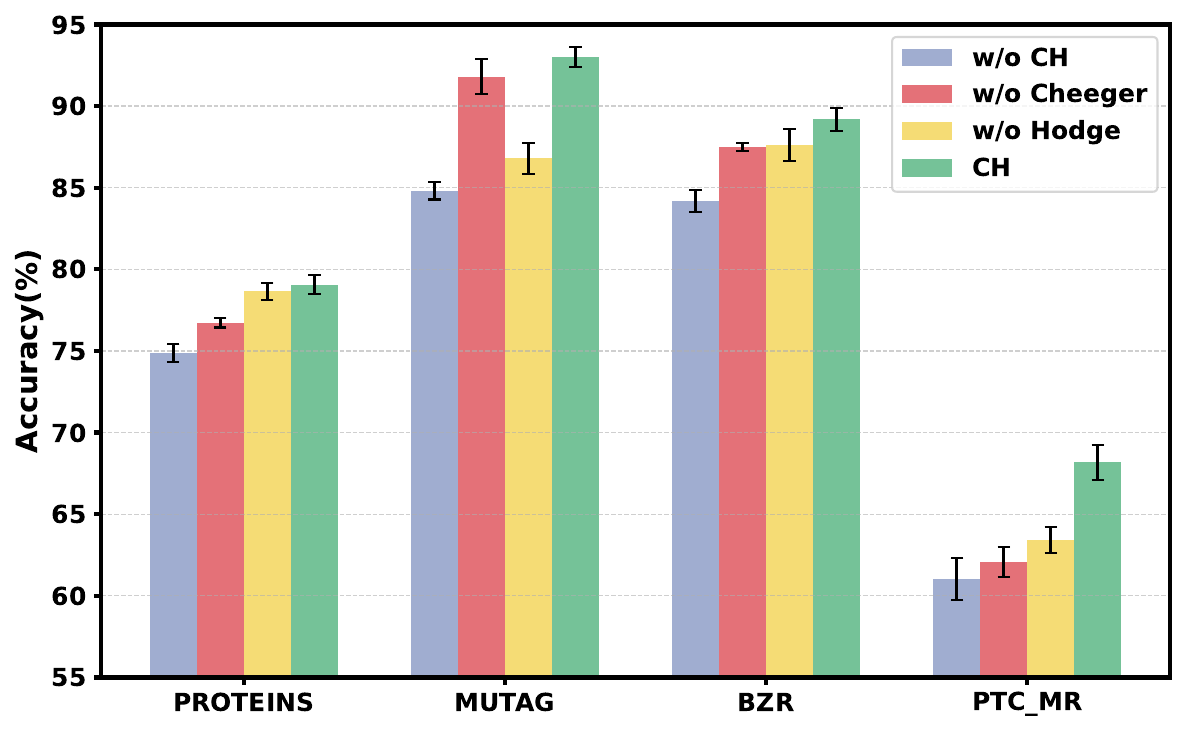}
  \caption{Ablation study comparing CHCL with three ablated variants that remove different components of the proposed framework.}
  \label{fig:ablation}
\end{figure}

As shown in Fig.~\ref{fig:ablation}, removing either the Cheeger module or the Hodge module consistently weakens performance, while removing both leads to the most pronounced degradation. These results indicate that both modules play essential roles in CHCL and that neither can be fully replaced by the other. More importantly, the larger performance drop observed when both are removed suggests that the two modules are complementary: rather than contributing independently, they jointly enhance the model’s ability to capture informative structural characteristics of graphs.

A more detailed comparison across datasets further reveals that the two modules contribute in different yet cooperative ways. The Cheeger module mainly strengthens the model’s sensitivity to global connectivity and structural separability, while the Hodge module provides higher-order topological cues that are difficult to capture from local patterns alone. Their relative importance varies across datasets, reflecting differences in the dominant structural semantics required for discrimination. On some datasets, performance depends more heavily on the global partition and connectivity information emphasized by the Cheeger module; on others, higher-order topological relationships encoded by the Hodge module play a more critical role. Nevertheless, the full model consistently achieves the strongest performance, showing that combining these two forms of structural information leads to more expressive graph representations and is key to the effectiveness of CHCL.

% \begin{figure*}[t]
%     \centering

%     \begin{subfigure}[t]{0.32\textwidth}
%         \centering
%         \includegraphics[width=\linewidth]{figtures/ablation.pdf}
%         \caption{Component ablation.}
%         \label{fig:ablation_component}
%     \end{subfigure}
%     \hfill
%     \begin{subfigure}[t]{0.32\textwidth}
%         \centering
%         \includegraphics[width=\linewidth]{figtures/cheeger_strategy_bar.pdf}
%         \caption{Cheeger construction strategies.}
%         \label{fig:ablation_cheeger_strategy}
%     \end{subfigure}
%     \hfill
%     \begin{subfigure}[t]{0.32\textwidth}
%         \centering
%         \includegraphics[width=\linewidth]{figtures/hodge_strategy_bar.pdf}
%         \caption{Hodge construction strategies.}
%         \label{fig:ablation_hodge_strategy}
%     \end{subfigure}

%     \caption{Ablation studies of CHCL. The component ablation evaluates the contribution of different modules, while the strategy-level ablations compare different construction strategies for the Cheeger and Hodge components.}
%     \label{fig:ablation}
% \end{figure*}

\subsubsection{Cheeger Dimension and Hodge Dimension}

\begin{figure*}[t]
    \centering

    \begin{subfigure}{0.49\textwidth}
        \centering
        \includegraphics[width=\linewidth]{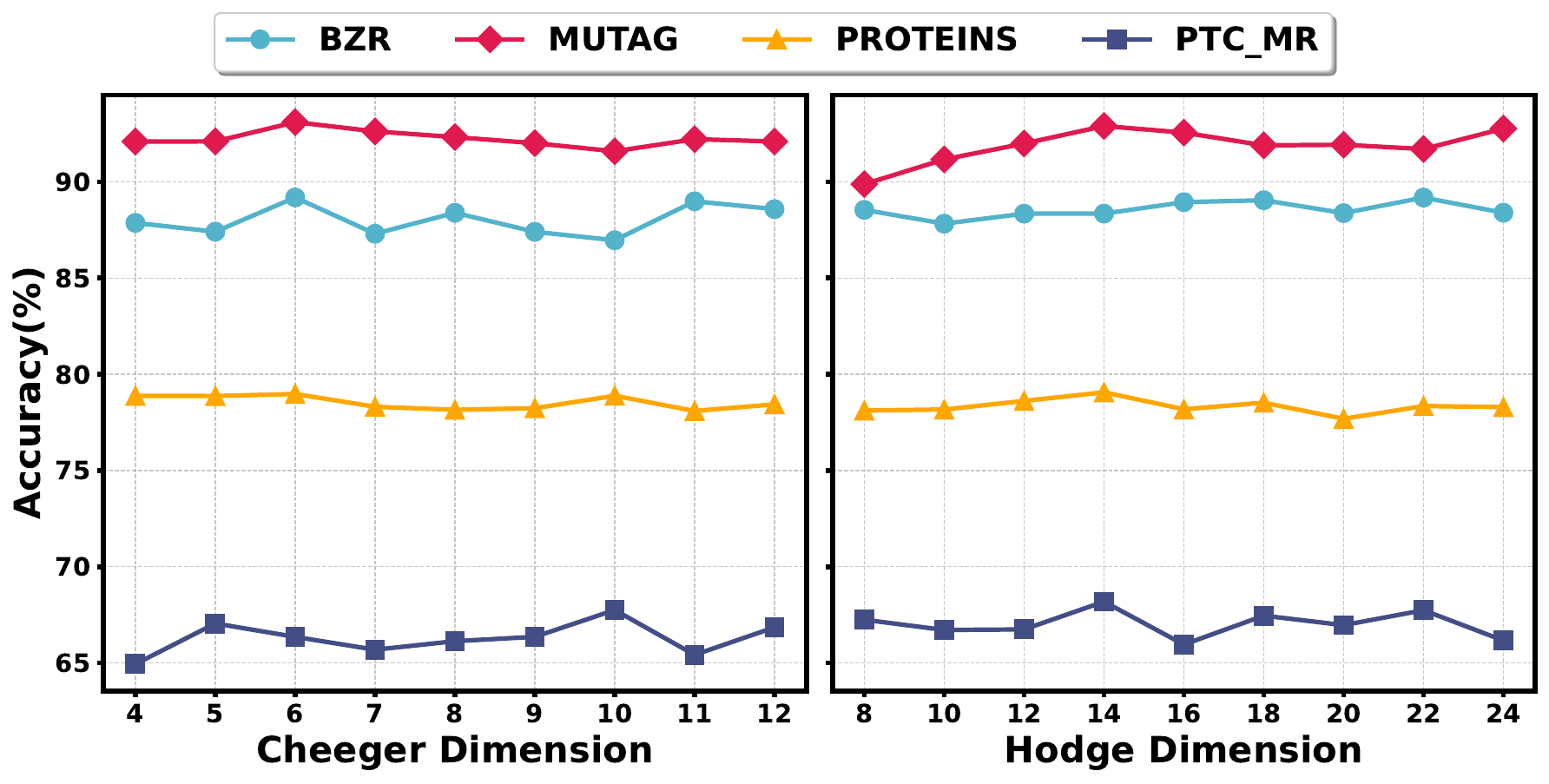}
        \caption{Sensitivity analysis of signature dimensions in the Cheeger and Hodge components.}
        \label{fig:dim}
    \end{subfigure}
    \hfill
    \begin{subfigure}{0.49\textwidth}
        \centering
        \includegraphics[width=\linewidth]{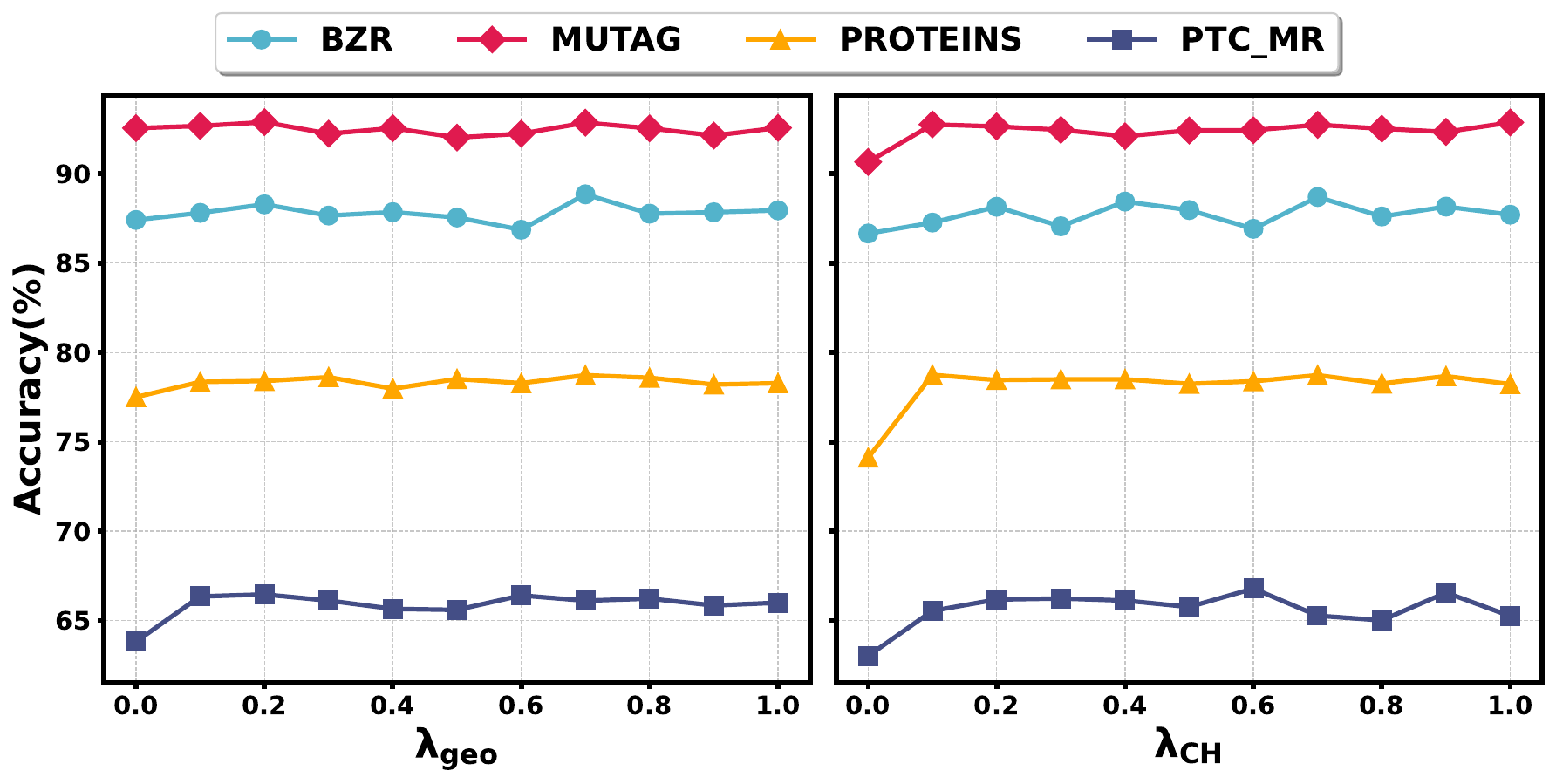}
        \caption{Sensitivity analysis of loss coefficients $\lambda_{geo}$ and $\lambda_{CH}$ for the geometry branch and the Cheeger--Hodge joint signature.}
        \label{fig:lambda}
    \end{subfigure}

    \caption{Sensitivity analysis of key hyperparameters in CHCL.}
    \label{fig:ablation_sensitivity}
\end{figure*}

% \begin{figure*}
%     \centering
%     \includegraphics[width=\linewidth]{figtures/combined_drop_edges_mask_feats_curve2.pdf}
%     \caption{Performance comparison of CHCL and baseline methods under varying levels of edge dropping and feature masking perturbations.}
%     \label{fig:edge_feat}
% \end{figure*}

\begin{figure*}[htbp]
    \centering
    \begin{subfigure}[b]{0.48\textwidth}
        \centering
        \includegraphics[width=\linewidth]{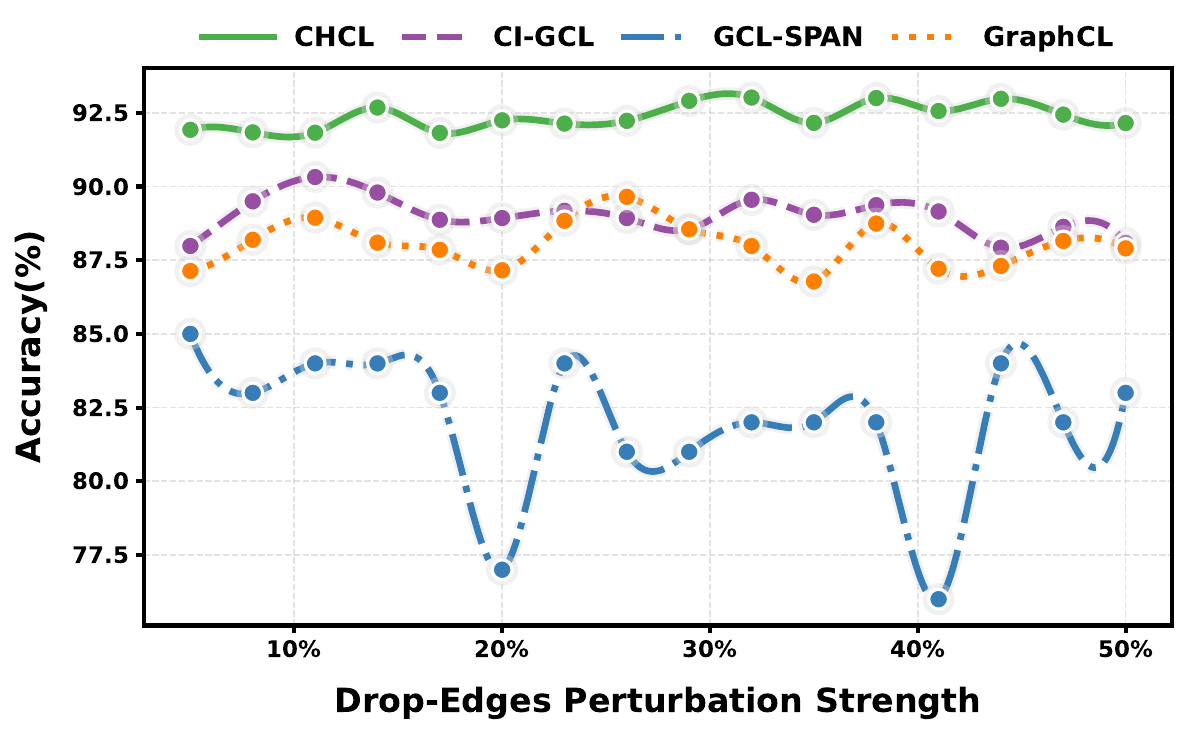}
        \caption{Robustness analysis under edge dropping.}
        \label{fig:edge_drop}
    \end{subfigure}
    \hfill
    \begin{subfigure}[b]{0.48\textwidth}
        \centering
        \includegraphics[width=\linewidth]{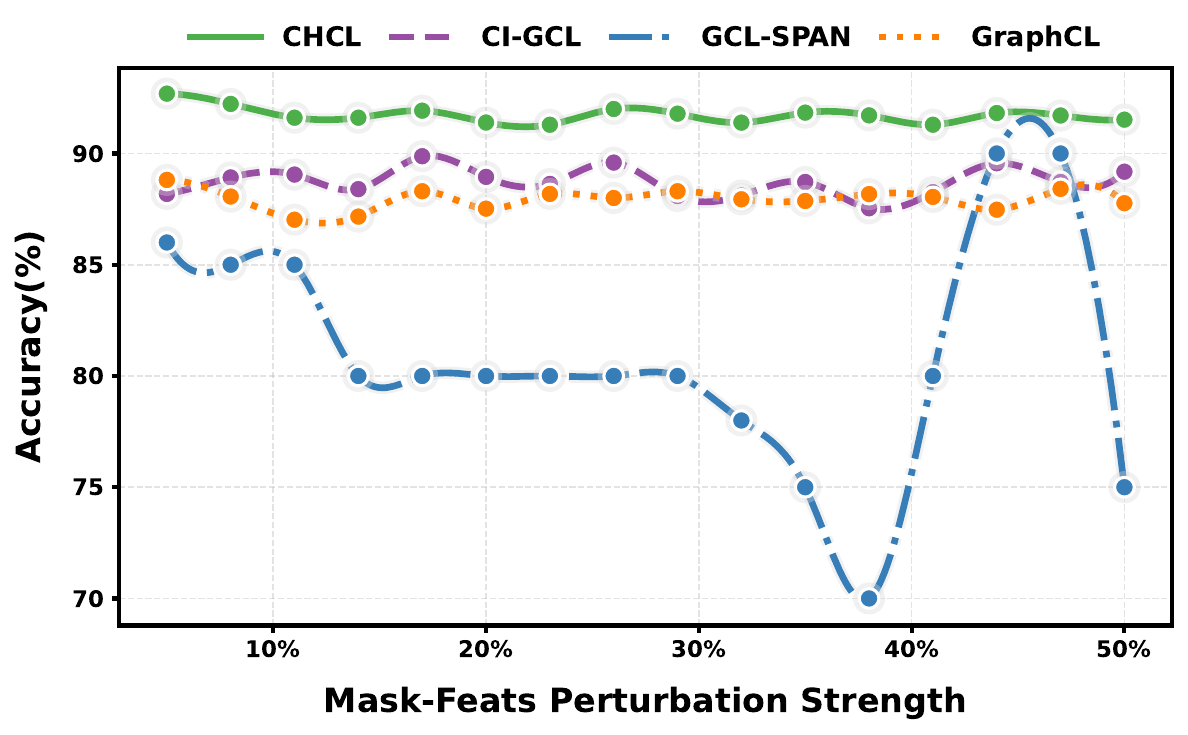}
        \caption{Robustness analysis under feature masking.}
        \label{fig:feat_mask}
    \end{subfigure}
    \caption{Performance comparison of CHCL and baseline methods under varying levels of edge dropping and feature masking perturbations.}
    \label{fig:edge_feat}
\end{figure*}

To understand how the signature dimensions of the Cheeger and Hodge components affect performance, we examine the sensitivity of CHCL to the dimensions of these two components. Specifically, we conduct experiments on the signature dimensions of the Cheeger and Hodge components: (i) we fix the Hodge signature dimension at $16$ and vary the Cheeger dimension from $4$ to $12$ with a step size of $1$; and (ii) we fix the Cheeger signature dimension at $8$ and vary the Hodge dimension from $8$ to $24$ with a step size of $2$.

Experimental results in Fig.~\ref{fig:dim} show that neither component requires a large signature dimension to be effective. Performance typically peaks at moderate dimensionalities and quickly enters a plateau. Across most datasets, increasing the dimension from a small value to a moderate range yields noticeable gains, whereas further enlarging the dimension brings rapidly diminishing returns and may even lead to slight degradation. Importantly, the curves exhibit a relatively broad near-optimal region around the peak, where performance varies only marginally across different dimensional choices. As can be seen from Fig.~\ref{fig:dim}, the overall optimum for the Cheeger component is usually attained with $d_{Cheeger}\in[6,10]$, while the Hodge component tends to perform best around $d_{Hodge}\approx 14$, suggesting that the structural signals captured by these components have relatively low effective complexity and can be represented compactly. Taken together, these findings indicate that CHCL is robust to the choice of signature dimensions, and adopting moderate dimensions is sufficient to achieve near-optimal performance in practice.

% \begin{figure}[t]
%     \centering
%     \begin{minipage}[b]{0.48\linewidth}
%         \centering
%         \includegraphics[width=\linewidth]{figtures/cheeger.pdf}
%     \end{minipage}
%     \hfill
%     \begin{minipage}[b]{0.48\linewidth}
%         \centering
%         \includegraphics[width=\linewidth]{figtures/hodge.pdf}
%     \end{minipage}
%     \caption{Ablation study on the feature dimensions of the Cheeger and Hodge components.}
%     \label{fig:dim}
% \end{figure}

% \begin{figure}
%     \centering
%     \includegraphics[width=\linewidth]{figtures/cheeger_hodge.pdf}
%     \caption{Ablation study on the feature dimensions of the Cheeger and Hodge components.}
%     \label{fig:dim}
% \end{figure}
\iffalse
\begin{figure}
    \centering

    \begin{subfigure}{\linewidth}
        \centering
        \includegraphics[width=\linewidth]{figtures/cheeger_hodge.pdf}
        \caption{Sensitivity analysis of signature dimensions in the Cheeger and Hodge components.}
        \label{fig:dim}
    \end{subfigure}

    \vspace{0.5em}

    \begin{subfigure}{\linewidth}
        \centering
        \includegraphics[width=\linewidth]{figtures/geo_joint.pdf}
        \caption{Sensitivity analysis of loss coefficients $\lambda_{geo}$ and $\lambda_{CH}$ for the geometry branch and the Cheeger--Hodge joint signature.}
        \label{fig:lambda}
    \end{subfigure}

    \caption{Sensitivity analysis of key hyperparameters in CHCL.}
    \label{fig:ablation_sensitivity}
\end{figure}
\fi

\subsubsection{Loss Coefficients $\lambda_{\text{geo}}$ and $\lambda_{\text{CH}}$}
To elucidate the respective roles of $\mathcal{L}_{\text{geo}}$ and $\mathcal{L}_{\text{CH}}$ during training, we investigate the sensitivity of CHCL to the loss weights $\lambda_{\text{geo}}$ and $\lambda_{\text{CH}}$. Specifically, we fix one coefficient at $1.0$ and vary the other in $(0,1)$ with a step size of $0.1$. This study aims to assess (i) whether both objectives are necessary and (ii) how robust the model is to the choice of coefficients.

As shown in Fig.~\ref{fig:lambda}, we observe that setting $\lambda_{\text{CH}}=0$ consistently leads to clear performance degradation across all four datasets, indicating that relying solely on the geometric contrastive constraint induced by random augmentations is insufficient to preserve the structural semantics required for graph discrimination. Meanwhile, incorporating $\mathcal{L}_{\text{geo}}$ further improves performance, suggesting that both $\mathcal{L}_{\text{geo}}$ and $\mathcal{L}_{\text{CH}}$ are necessary. Notably, when both $\lambda_{\text{geo}}$ and $\lambda_{\text{CH}}$ are nonzero, performance becomes largely insensitive to the exact choice of coefficients: over a wide range of $\lambda_{\text{geo}}$ and $\lambda_{\text{CH}}$, the curves exhibit a pronounced plateau with only minor fluctuations. Overall, our sensitivity analysis of the loss coefficients shows that Cheeger--Hodge signatures serve as essential semantic anchors for graphs, and coupling them with geometric alignment further improves model performance.

\subsection{Robustness to Augmentation Perturbations}

To validate the effectiveness of the Cheeger--Hodge joint signature as a semantic invariance anchor for graphs, we conduct controlled robustness experiments by perturbing the input graphs with edge dropping and feature masking. The perturbation strength is varied from 0.05 to 0.50; at each level, we randomly drop edges or mask node features according to the specified strength. To comprehensively compare the model's performance under these perturbations, we selected three representative contrastive learning baseline methods: CI-GCL, GCL-SPAN, and GraphCL.

As shown in Fig.~\ref{fig:edge_drop}, CHCL remains consistently competitive under edge dropping across a wide range of perturbation strengths and exhibits stronger stability than the baseline methods. These results suggest that the Cheeger--Hodge joint signature provides a stable semantic invariance anchor throughout the augmentation process. By grounding representation learning in global structural organization and higher-order topological information, CHCL is less affected by random perturbations that alter local structural patterns. As a result, it can preserve intrinsic graph semantics more effectively and avoid the instability commonly induced by aggressive augmentations in conventional contrastive learning.

A similar trend is observed under feature masking, as shown in Fig.~\ref{fig:feat_mask}. CHCL remains consistently competitive across different masking ratios, demonstrating strong robustness to attribute-level corruption. Since feature masking perturbs node attributes but does not change the graph structure, the Cheeger--Hodge signatures remain stable and continue to serve as reliable semantic guidance during learning. This shows that incorporating structure-derived invariance not only improves resilience to structural perturbations, but also strengthens robustness when node features are incomplete or heavily corrupted.

\subsection{Interpretability}
\begin{figure*}[htbp]
    \centering

    % 第一行左图
    \begin{subfigure}[t]{0.45\textwidth}
        \centering
        \includegraphics[width=\linewidth]{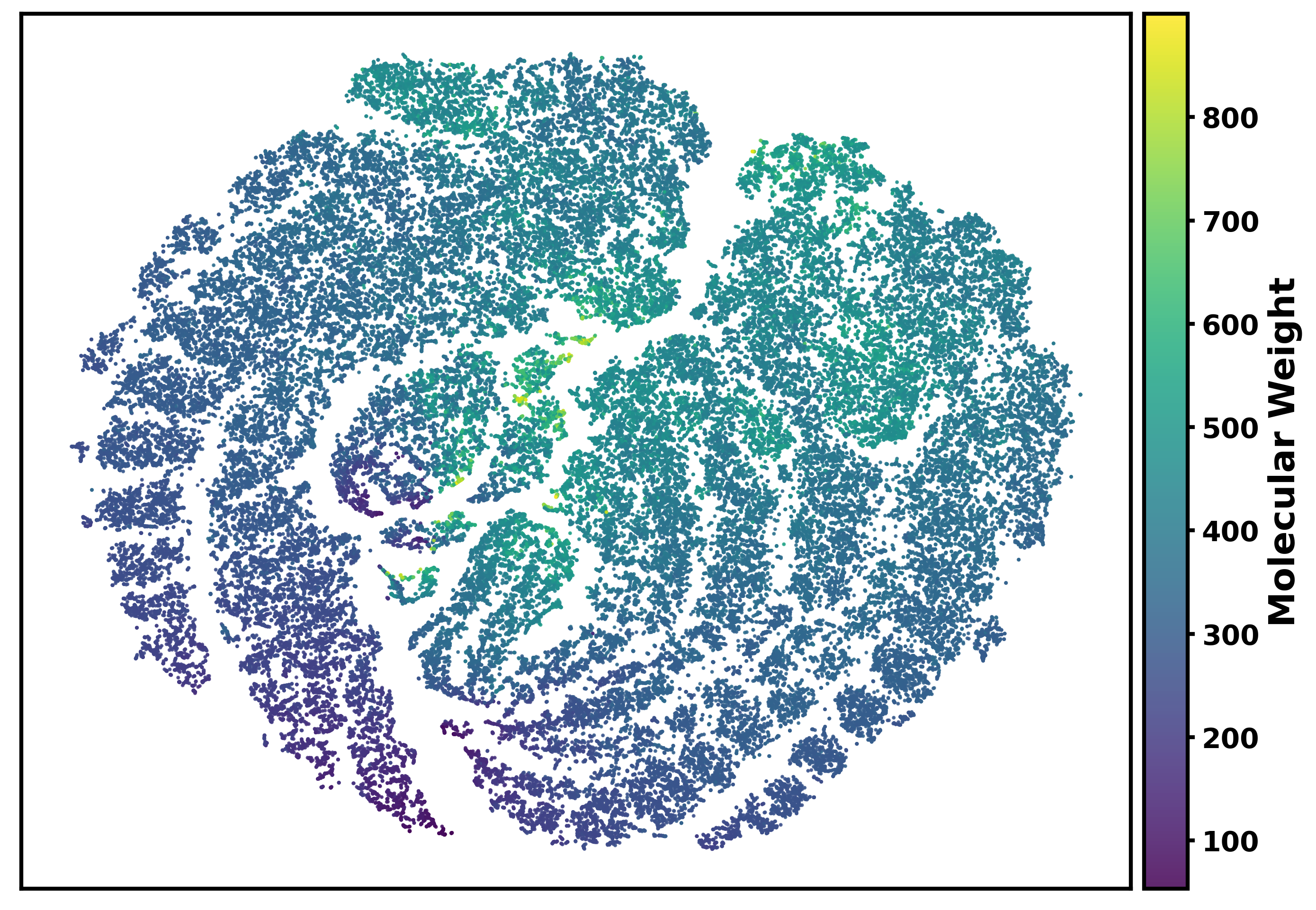}
        \caption{Global organization of pretrained molecular representations on ChEMBL, visualized by t-SNE and colored by molecular weight.}
        \label{fig:visual_a}
    \end{subfigure}
    \hfill
    % 第一行右图
    \begin{subfigure}[t]{0.45\textwidth}
        \centering
        \makebox[\linewidth][c]{%
            \rlap{\raisebox{-2mm}[0pt][0pt]{%
                \includegraphics[width=\linewidth]{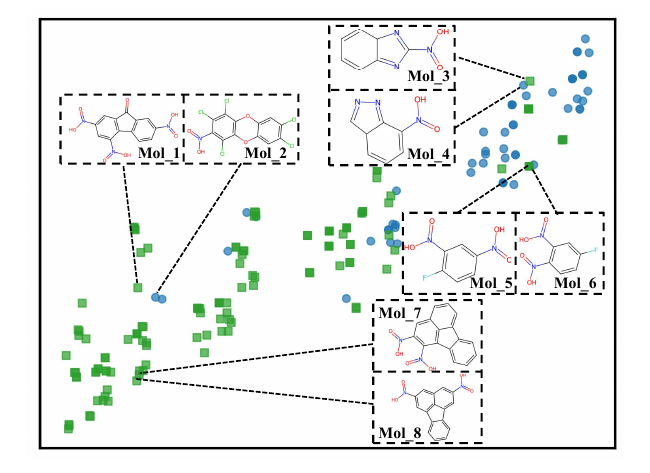}%
            }}%
            \phantom{\includegraphics[width=\linewidth]{figtures/MUTAG.pdf}}%
        }
        \caption{Label-relevant structural separation in the Cheeger--Hodge joint-signature space on MUTAG.}
        \label{fig:visual_b}
    \end{subfigure}

    \caption{Interpretability of the structural semantics learned by CHCL.}
    \label{fig:visual}
\end{figure*}
To further examine whether CHCL learns interpretable structural semantics, we visualize the learned representations from four perspectives: (i) whether the learned representation captures the overall molecular organization and structural scale of molecules; (ii) whether the Cheeger--Hodge signatures capture meaningful structural semantics relevant to graph labels.

Figure~\ref{fig:visual_a} presents a t-SNE visualization of the pretrained representations on the ChEMBL dataset, where samples are colored by molecular weight. While molecular weight is not a complete descriptor of molecular structure, it serves as a coarse indicator of overall molecular scale, with higher values typically corresponding to larger atomic compositions and more extended scaffolds. The smooth color transition and coherent regional clustering in the embedding space suggest that CHCL captures variation in molecular scale together with coarse-grained structural organization. This observation indicates that, even in the absence of label supervision, the pretrained model learns a representation space with meaningful global discriminability, grouping molecules with similar overall structural characteristics into nearby regions.

Figure~\ref{fig:visual_b} further examines the Cheeger--Hodge signatures learned from unsupervised training on MUTAG, where colors and marker shapes indicate different classes. The visualization shows that molecules from different classes exhibit a clear tendency to separate in the Cheeger--Hodge signature space, indicating that the learned signatures encode structural semantics that are highly predictive of graph labels. Meanwhile, a small number of molecules remain not well separated. We further highlight three groups of such molecules and one pair of closely located molecules, and conduct a comparative analysis of their chemical structures. We find that these molecules often share highly similar cyclic substructures, which naturally results in close proximity within the Cheeger--Hodge embedding space. This observation also explains the gains from geometric alignment in our ablation studies: Cheeger--Hodge signatures tend to cluster molecules with similar higher-order topology, while geometric-view alignment injects complementary cues about local structure, helping to distinguish molecules that share cyclic substructures but differ in subtle details. Overall, this visualization provides evidence that the learned Cheeger--Hodge signatures capture label-relevant higher-order structural semantics, thereby offering an interpretable structural basis for the discriminative power of CHCL.

Taken together, these perspectives consistently support the same conclusion: CHCL learns a representation space with meaningful semantic structure. At the global level, it organizes molecules according to interpretable property-related signals; at the discriminative level, Cheeger--Hodge signatures capture label-relevant structural semantics. These results indicate that the representation space learned by CHCL is not an arbitrary latent geometry, but a structurally and chemically grounded semantic space, providing a plausible explanation for the strong robustness and transferability of the model.

\section{Conclusion}
In this work, we introduced the Cheeger--Hodge joint signature as an intrinsic structural descriptor of graphs, which allows the learned representations to better preserve the graph's global structural properties and higher-order topological information. More importantly, our method admits a Lipschitz stability guarantee under graph perturbations, ensuring that the essential structural information of a graph can be stably retained during augmentation. As a result, CHCL provides a more rigorous notion of structural robustness grounded in mathematical stability, rather than relying solely on the empirical resilience often observed in existing methods. Our experimental results demonstrate that leveraging Cheeger--Hodge signatures not only improves model performance but also provides greater robustness, establishing a powerful framework for graph contrastive learning.

% \section*{Acknowledgment}


\ifCLASSOPTIONcaptionsoff
  \newpage
\fi

\end{document}